\documentclass[10pt,twocolumn,letterpaper]{article}

\usepackage{iccv}
\usepackage{times}
\usepackage{epsfig}
\usepackage{graphicx}
\usepackage{amsmath}
\usepackage{amssymb}
\usepackage[numbers]{natbib}
\usepackage[T1]{fontenc}
\usepackage{multirow}

\usepackage[colorlinks,linkcolor=blue]{hyperref}
\iccvfinalcopy 

\def\iccvPaperID{8} 

\ificcvfinal\fi
\usepackage{url}

\begin{document}

\title{EdgeFlow: Achieving Practical Interactive Segmentation \\ with Edge-Guided Flow}

\author{Yuying Hao$^1$, Yi Liu$^1$, Zewu Wu$^1$, Lin Han$^{2*}$, Yizhou Chen$^{3*}$, Guowei Chen$^1$ \\ Lutao Chu$^1$, Shiyu Tang$^1$, Zhiliang Yu$^1$, Zeyu Chen$^1$, Baohua Lai$^1$ \\

{$^1$Baidu, Inc. \quad $^2$NYU \quad $^3$CQJTU}

}

\maketitle
\ificcvfinal\thispagestyle{empty}\fi
\renewcommand{\thefootnote}{\fnsymbol{footnote}} 
\footnotetext[1]{PaddlePaddle Developers Experts (PPDE)} 

\begin{abstract}

High-quality training data play a key role in image segmentation tasks. Usually, pixel-level annotations are expensive, laborious and time-consuming for the large volume of training data. To reduce labelling cost and improve segmentation quality, interactive segmentation methods have been proposed, which provide the result with just a few clicks. However, their performance does not meet the requirements of practical segmentation tasks in terms of speed and accuracy. In this work, we propose EdgeFlow, a novel architecture that fully utilizes interactive information of user clicks with edge-guided flow. Our method achieves state-of-the-art performance without any post-processing or iterative optimization scheme. Comprehensive experiments on benchmarks also demonstrate the superiority of our method. In addition, with the proposed method, we develop an efficient interactive segmentation tool for practical data annotation tasks. The source code and tool is avaliable at \href{https://github.com/PaddlePaddle/PaddleSeg}{https://github.com/PaddlePaddle/PaddleSeg}.

\end{abstract}


\section{Introduction}

Deep learning has seen tremendous success in computer vision areas, such as image recognition~\cite{recognation1,recognition2}, object detection~\cite{detection1, detection2} and image segmentation~\cite{deeplab, fcn, paddleseg}. In order to learn powerful abstraction, large volumes of labelled image data are usually essential for the model training process. As the amount of data increasing, the cost of manual annotation grows rapidly, especially when it comes to pixel-level segmentation tasks. Although semi-supervised or even unsupervised algorithms have been proposed to relieve label dependence, there is a great gap in accuracy between them and full supervision. 

Therefore, interactive segmentation appears to be an attractive and efficient way, which allows the human annotators to quickly extract the object-of-interest~\cite{zhang2020interactive}. Unlike model-centric methods, interactive segmentation methods take interactive information into account. Therefore, they simplify the annotation process and improve the quality progressively. In general, interactive information could be various inputs, such as scribbles~\cite{scribbles}, clicks~\cite{ritm,fca,f-brs}, bounding boxes~\cite{bbox} and so on. 
\begin{figure}[pt]
	\begin{center}
		\begin{tabular}{cc}
			
			\includegraphics[width=0.44\linewidth,height=2.4cm]{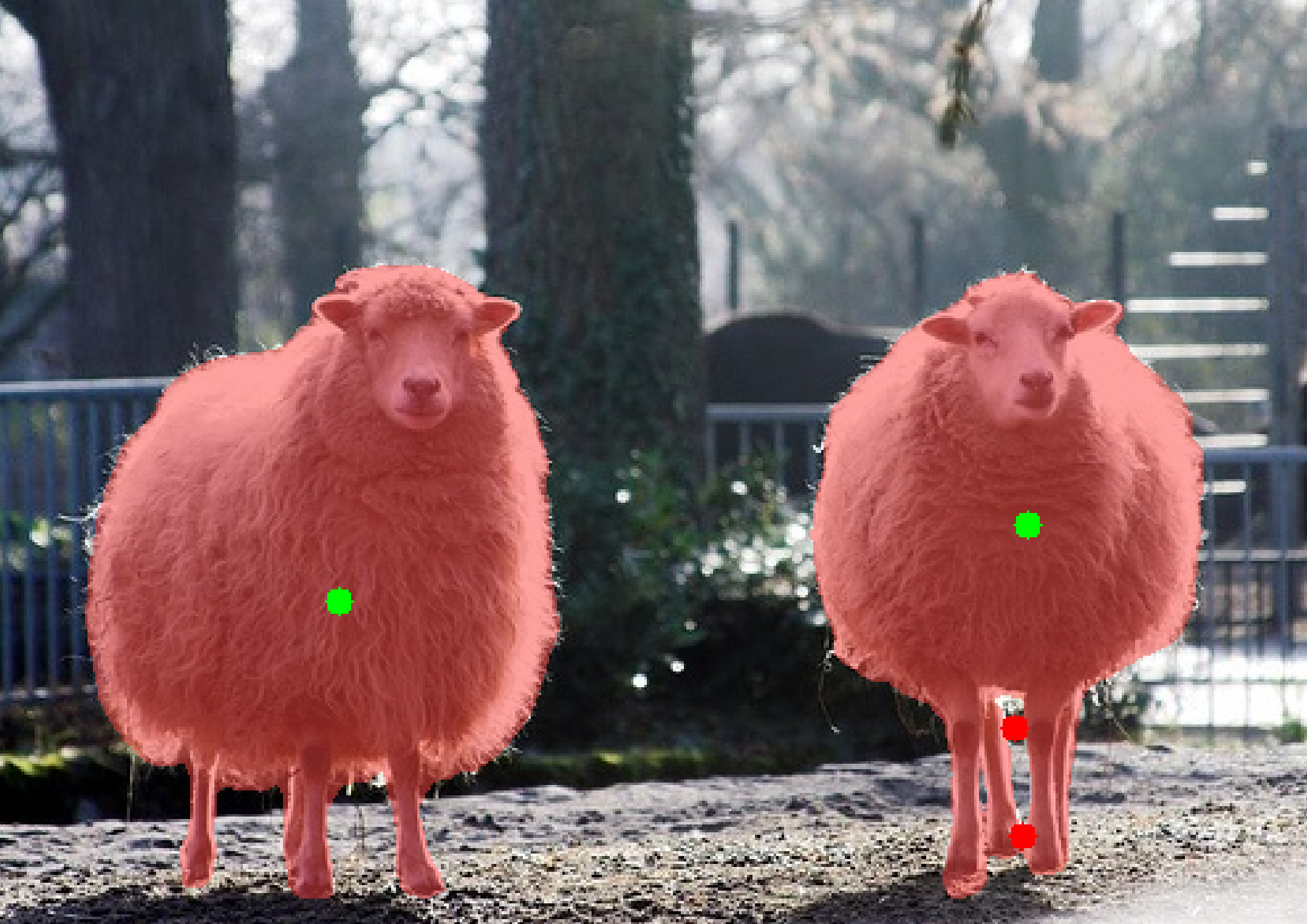}&
			\includegraphics[width=0.44\linewidth,height=2.4cm]{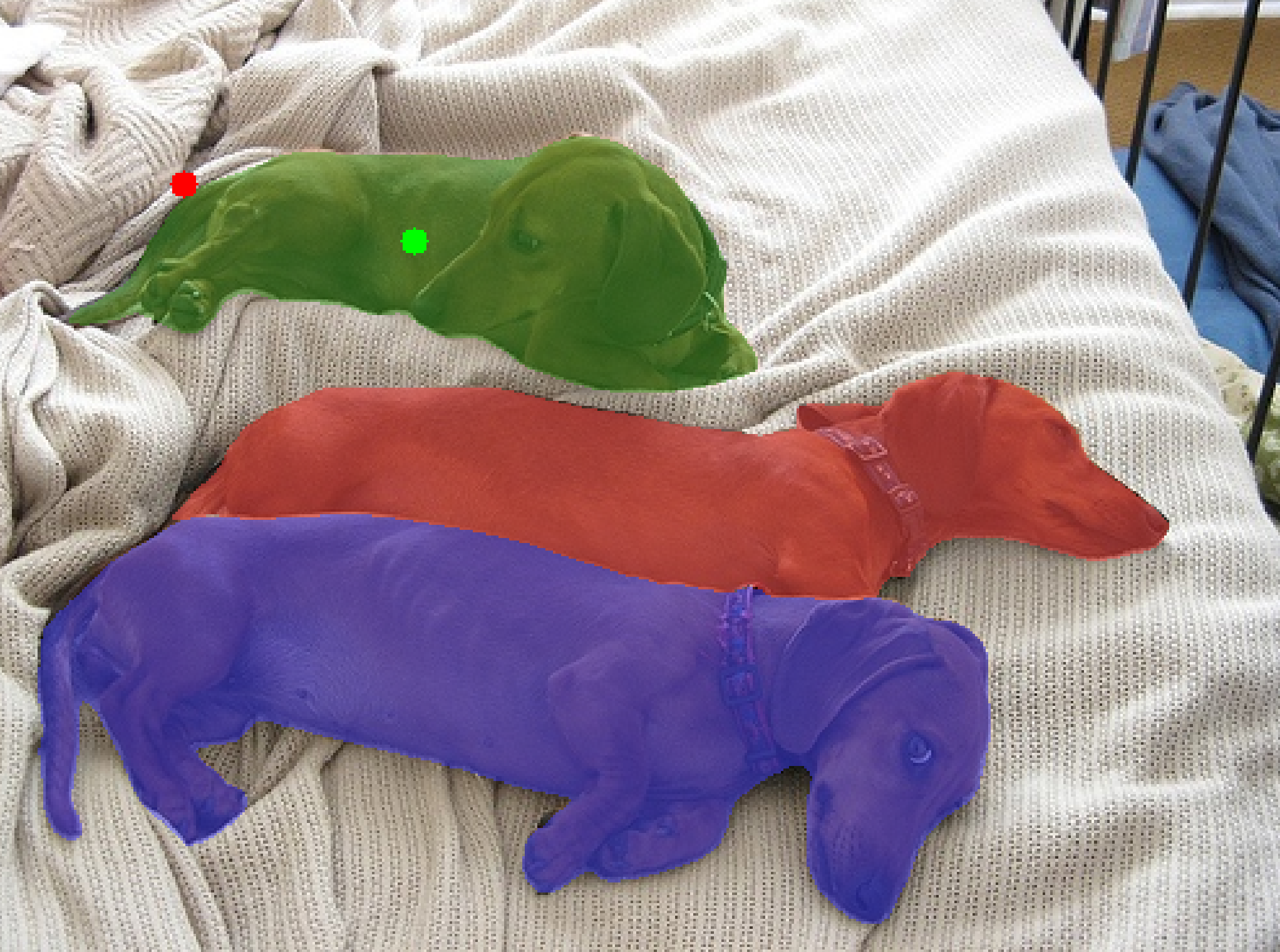}
		\end{tabular}
	\end{center}
\label{fig:show_image}
\caption{Examples of interactive clicks. Green dots denote positive clicks, red dots denote negative clicks. Best viewed in colors.} 
\vspace{-0.15in}
\end{figure}

\begin{figure*}[tph]
	\begin{center}
		\includegraphics[width=0.9\linewidth]{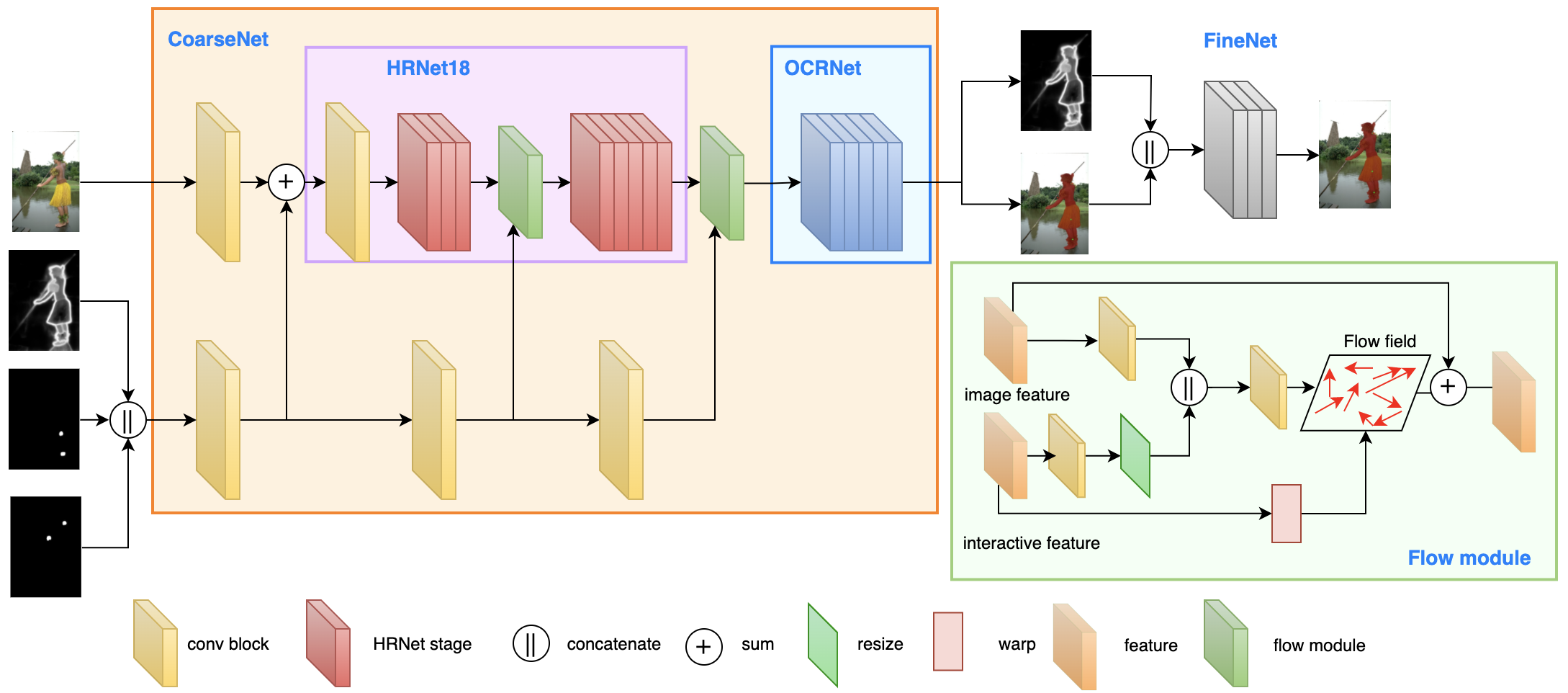}
	\end{center}
	\caption{Overview of the EdgeFlow architecture. We design a coarse-to-fine network including CoarseNet and FineNet. For CoarseNet, We utilize HRNet-18+OCR as the base segmentation model and append the edge-guided flow to deal with interactive information. For FineNet, we utilize three atrous convolution blocks to refine the coarse masks.}
	\label{fig:structure}
	\vspace{-0.15in}
\end{figure*}
The characteristics of the different interactions have been studied well by previous works, where click-based methods are the most promising, because they provide sufficient selected object information with minimal interaction time. In practice, click-based methods usually use two types of user clicks, i.e. positive clicks and negative clicks. Positive clicks aim to emphasize the target object (foreground) and negative clicks isolate non-target areas (background). Usually, such methods only need a few clicks to complete an object segmentation task, which is shown in Fig.~\ref{fig:show_image}.  

In recent years, there are a couple of works~\cite{eccv2020, brs,f-brs} on click-based interactive segmentation, in which deep learning methods surpass traditional ones in terms of accuracy. However, most of them require extra post-processing in the evaluation process, which is time-consuming in practice. More recently, end-to-end interactive algorithms~\cite{fca, ritm} were proposed to speed up the clicking interaction, but they have common problems. 
The clicks are the only input to the first layers, so that the specific spatial and semantic information would be diluted through early layers. The other problem is that the relation of consecutive clicks is not modelling properly, resulting in unstable annotations, e.g. the segmentation annotation dramatically changes between two consecutive clicks.

In this work, we propose a novel interactive segmentation architecture that fully utilizes clicks of users and the relation of consecutive clicks. To enhance the interactive information, the feature of user clicks are embedded into both early and late layers and image features are efficiently integrated with an early-late fusion strategy. To establish the relationship between two consecutive clicks, the edge mask generated by the previous clicks is taken as an input together with the current click. It improves the stability of segmentation results significantly. In addition, we adopt a coarse-to-fine network design to further obtain the fine-grained segmentation. The comprehensive evaluations show our state-of-the-art performance on well-known benchmarks.

Furthermore, upon the proposed interactive model, we develop an efficient interactive segmentation tool for practical segmentation tasks, e.g. image labelling. The tool not only generates the segmentation mask, but also allows the user to adjust the polygon vertexes of the mask to further improve accuracy. Thus, the tool provides flexible options for annotation accuracy according to different practical tasks.

Our contributions are summarized as follow:

\begin{itemize}
  \item We propose a novel interactive architecture that fully utilizes interactive and image information with the early-late fusion. The enhancement of interactive clicks prevents feature dilution over the network and then enables it to respond to the clicks efficiently. 
  \item We utilize the object edges produced by network to improve the segmentation stability. With the coarse-to-fine network design, comprehensive experiments show our method achieves state-of-the-art performance on several benchmarks.
  \item We develop an efficient interactive segmentation tool that supports interactive annotation and polygon frame editing. Our tool also supports multi-scenes and various labelling formats.
\end{itemize}

\begin{figure*}[th]
	\setlength{\abovecaptionskip}{0.cm}
	\setlength{\belowcaptionskip}{-0.cm}
	\begin{center}
		\begin{tabular}{cccccc}
			
			\includegraphics[width=0.14\linewidth]{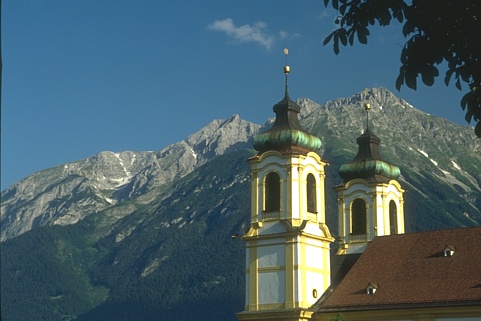}&
			\includegraphics[width=0.14\linewidth]{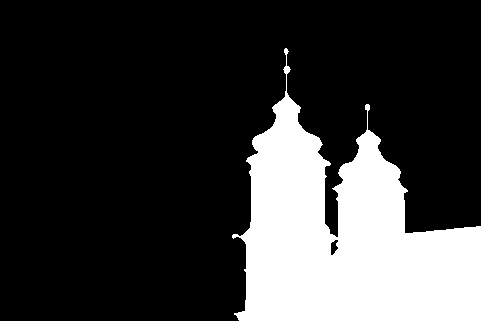}&
			\includegraphics[width=0.14\linewidth]{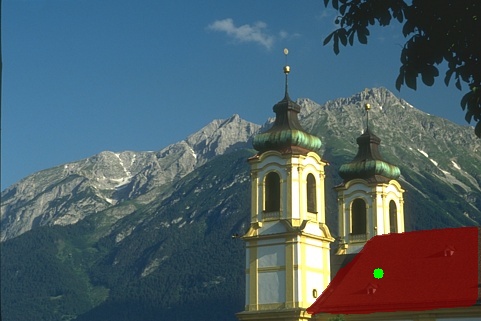}&
			\includegraphics[width=0.14\linewidth]{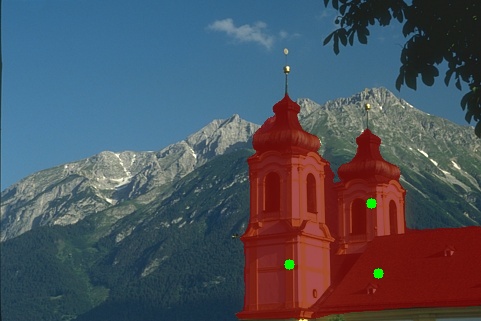}&
			\includegraphics[width=0.14\linewidth]{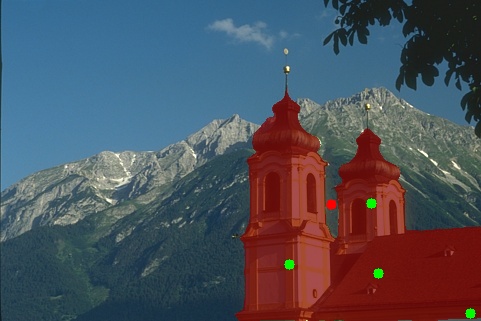}&
			\includegraphics[width=0.14\linewidth]{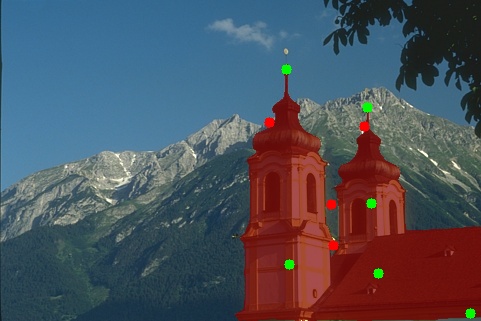} \\
			
			{\scriptsize image} &
			{\scriptsize GT Mask} &
			{\scriptsize 1 click IoU=31.6\%} &
			{\scriptsize 3 clicks IoU=89.7\%} &
			{\scriptsize 5 clicks IoU=96.9\%} &
			{\scriptsize 10 clicks IoU=97.0\%} \\

			\includegraphics[width=0.14\linewidth]{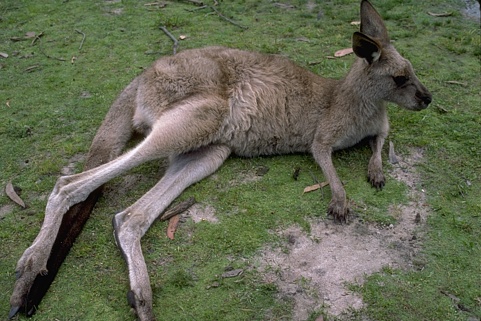}&
			\includegraphics[width=0.14\linewidth]{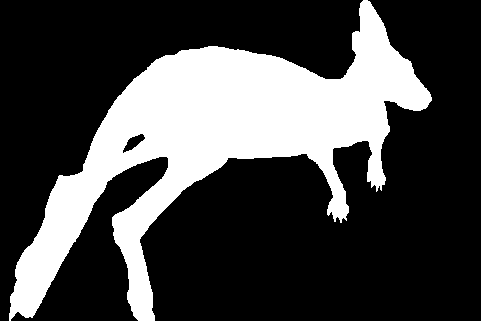}&
			\includegraphics[width=0.14\linewidth]{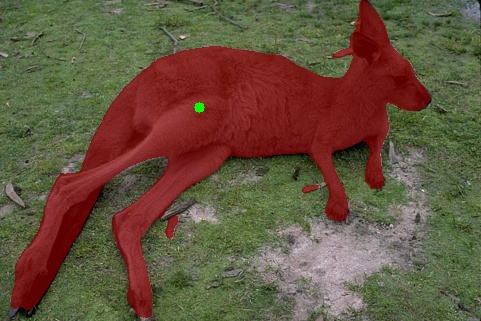}&
			\includegraphics[width=0.14\linewidth]{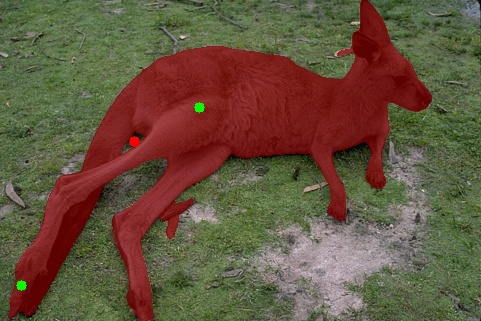}&
			\includegraphics[width=0.14\linewidth]{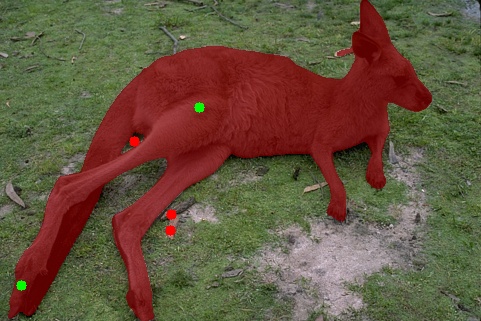}&
			\includegraphics[width=0.14\linewidth]{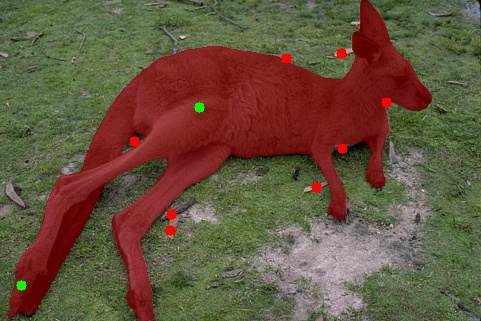} \\
			
			{\scriptsize image} &
			{\scriptsize GT Mask} &
			{\scriptsize 1 click IoU=94.0\%} &
			{\scriptsize 3 clicks IoU=94.9\%} &
			{\scriptsize 5 clicks IoU=96.2\%} &
			{\scriptsize 10 clicks IoU=96.5\%} \\

			\includegraphics[width=0.14\linewidth]{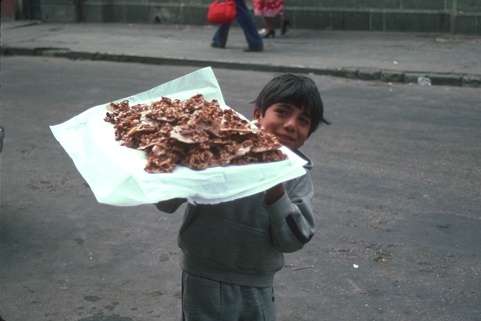}&
			\includegraphics[width=0.14\linewidth]{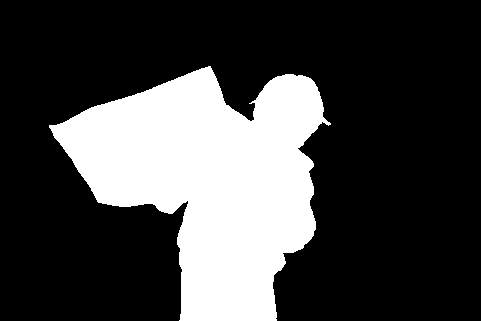}&
			\includegraphics[width=0.14\linewidth]{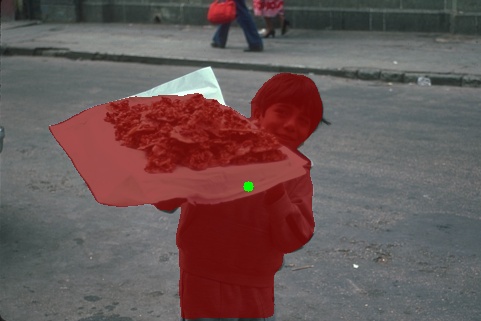}&
			\includegraphics[width=0.14\linewidth]{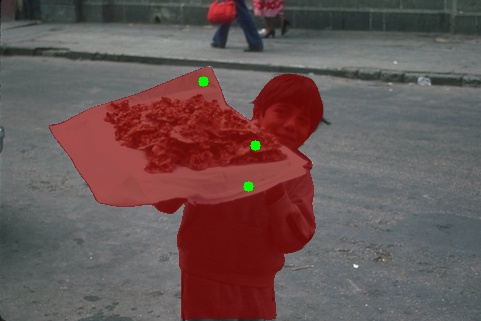}&
			\includegraphics[width=0.14\linewidth]{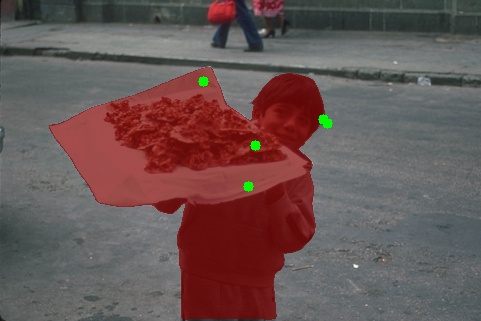}&
			\includegraphics[width=0.14\linewidth]{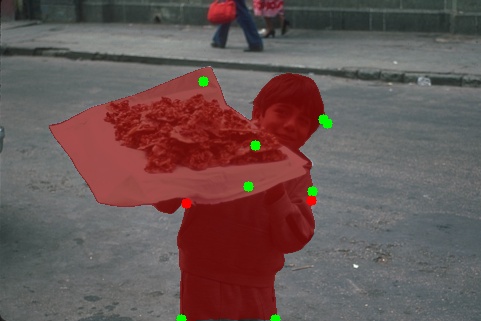} \\
			
			{\scriptsize image} &
			{\scriptsize GT Mask} &
			{\scriptsize 1 click IoU=93.4\%} &
			{\scriptsize 3 clicks IoU=98.4\%} &
			{\scriptsize 5 clicks IoU=98.5\%} &
			{\scriptsize 10 clicks IoU=98.6\%} \\
			
		\end{tabular}
		
	\end{center}
    \caption{Qualitative results of our method on Berkeley dataset. The first two columns represent original images and ground-truth masks, respectively. Other columns show segmentation results with the different numbers of clicks. Green dots denote positive clicks, red dots denote negative clicks.} 
    \label{fig:show_image1}
    \vspace{-0.15in}
\end{figure*}
\section{Related Works}
 
The interactive segmentation task aims to obtain an accurate mask of an object with minimal user interaction. Interactive information can be clicks, scratches, contours, bounding boxes, phrases and so on. According to the type of information modelling, there are two research branches for interactive image segmentation algorithms.

\textbf{Optimization based methods:} Most of them are traditional methods which are divided into four categories:  1) contour based methods~\cite{contour1,contour2, contour3, contour4, contour5},  2) graph-cut based methods~\cite{gc1, gc2, gc3}, 3) random walk based methods~\cite{rw1,rw2, rw3} and 4) region merging methods~\cite{rm1,rm2, rm3}.  
As a contour-based method, the active contours model (ACM)~\cite{acm1} constructs and optimizes an energy equation until the force on the closed curve decreases to zero. 
Graph-cut based methods (GC)~\cite{gc1, gc2} utilize min-cut/max-flow algorithm to minimize the energy function. 
Random walk based methods (RW)~\cite{rw1, rw2} construct an undirected graph by taking pixels as vertices and the relationship of neighbourhoods as edges. 
Region merging methods (RM) are initialized by user-interactive seeds and then gather the similar points and regions by homogeneity criterion. 

The traditional methods have common drawbacks. The generalization ability of the methods is poor, where they only work well on specific scenes. In addition, they are sensitive to initial interactive information and require high-quality interactions without noisy input.

\textbf{Deep learning based methods:} Xu~\etal~\cite{deep1} firstly introduced deep learning to solve interactive segmentation problems. They converted the interactive clicks to a distance map and then took the distance map together with the original image as an input to fine-tune FCN~\cite{fcn}. After that, Maninis~\etal~\cite{dextr} took extreme points of an object as interactive information and utilized DeepLabv2~\cite{deeplab} as a segmentation model. Optimization for activation has been applied on the back-propagating refinement scheme (BRS)~\cite{brs}, which corrected mislabeled pixels by employing the L-BFGS algorithm. However, the optimization is time-consuming. Soon after, Feature-BRS~\cite{f-brs} was proposed to improve the optimization scheme and speed up the interaction process. The mislabeled pixels would be rectified by auxiliary scales and biases which modify features from the middle of the network. Kontogianni~\etal~\cite{eccv2020} adjusted the mask of the target object by optimizing the model parameters at test time.  Sofiiuk~\etal~\cite{ritm} applied an iterative training process by employing previous mask outputs, which improved the segmentation accuracy.

However, these methods highly rely on post-processes or additional optimization schemes in the evaluation process, which requires extra time and computations. Besides, they are sensitive to a new click, resulting in unstable segmentation masks.

\section{Proposed Method}

In this work, we propose a novel interactive segmentation architecture that fully utilizes clicks and the relation of consecutive clicks. Compared with previous approaches that combine image and clicks together at the first layers, our proposed method 1) utilizes a separate branch to enhance the features of clicks, which enables the network to respond to interactive information better; 2) applies edge mask as prior information of the network, which stabilizes annotations.

\begin{figure*}[th]
	\setlength{\abovecaptionskip}{0.cm}
	\setlength{\belowcaptionskip}{-0.cm}
	\begin{center}
		\begin{tabular}{cccccc}
			\includegraphics[width=0.14\linewidth]{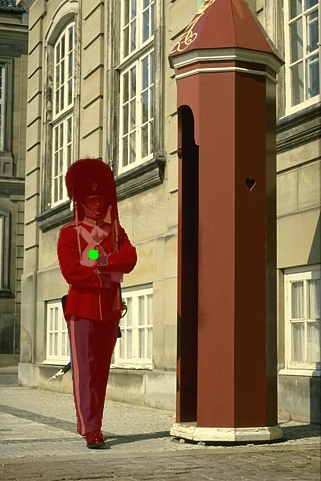}&
			\includegraphics[width=0.14\linewidth]{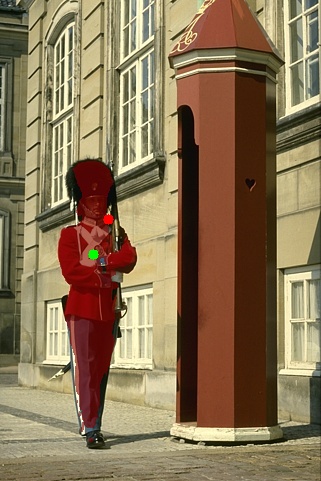}&
			\includegraphics[width=0.14\linewidth]{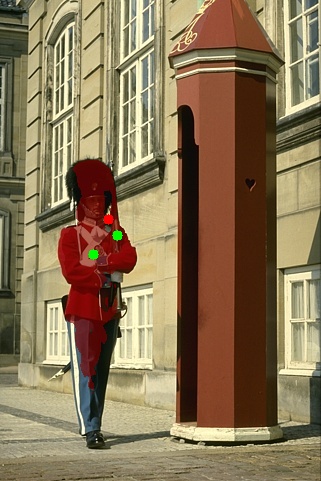}&
			\includegraphics[width=0.14\linewidth]{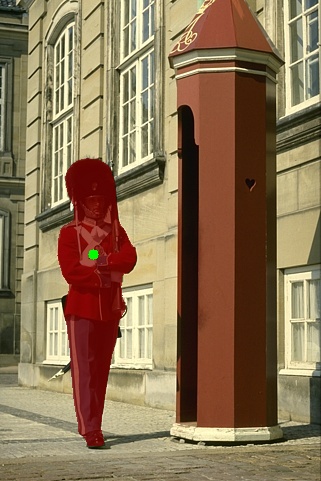}&
			\includegraphics[width=0.14\linewidth]{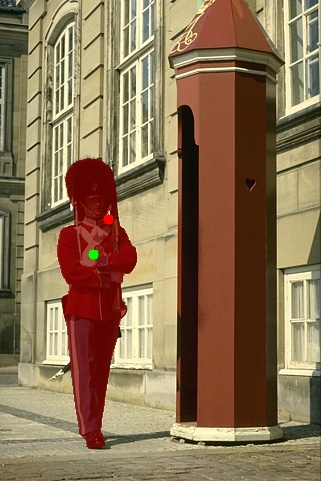}&
			\includegraphics[width=0.14\linewidth]{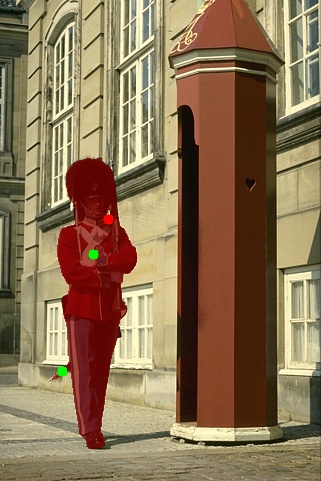} \\

			\includegraphics[width=0.14\linewidth]{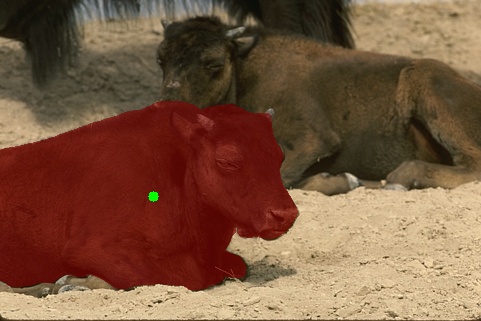}&
			\includegraphics[width=0.14\linewidth]{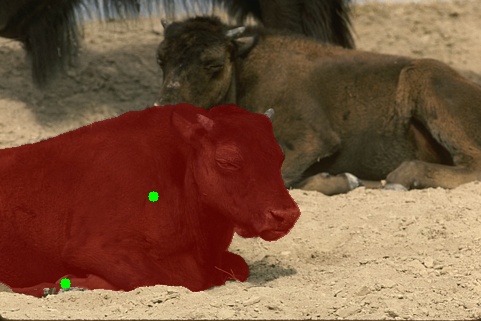}&
			\includegraphics[width=0.14\linewidth]{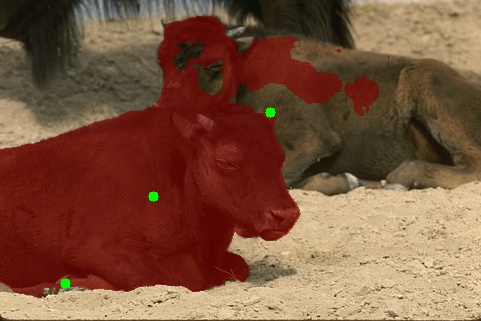}&
			\includegraphics[width=0.14\linewidth]{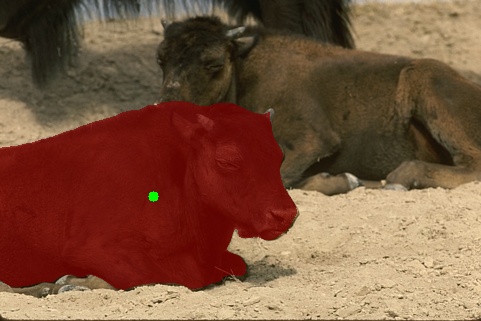}&
			\includegraphics[width=0.14\linewidth]{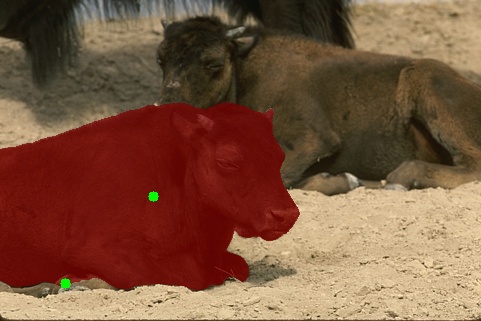}&
			\includegraphics[width=0.14\linewidth]{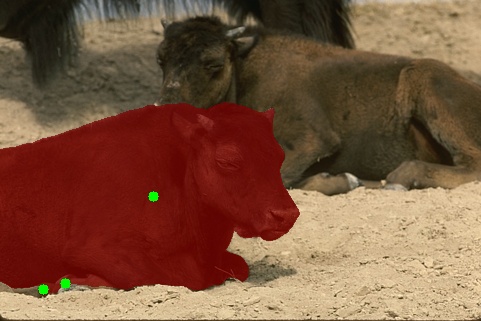} \\
			
			{\scriptsize baseline, 1 click} &
			{\scriptsize baseline, 2 clicks} &
			{\scriptsize baseline, 3 clicks} &
			{\scriptsize baseline+EF, 1 click} &
			{\scriptsize baseline+EF, 2 clicks} &
			{\scriptsize baseline+EF, 3 clicks} \\
			
		\end{tabular}
	\end{center}
    \caption{The qualitative results of edge-guided flow on Berkeley dataset.  The fist three columns show the output of baseline on different clicks. The last three columns show the output of baseline with edge-guided flow on different clicks. Edge-guided flow produces more stable segmentation results. } 
    \label{fig:final_base_vs_edge}

\end{figure*}

\begin{figure*}[th]
	\setlength{\abovecaptionskip}{0.cm}
	\setlength{\belowcaptionskip}{-0.cm}
	\begin{center}
		\begin{tabular}{cccccc}
			
			\includegraphics[width=0.14\linewidth,height=3.7cm]{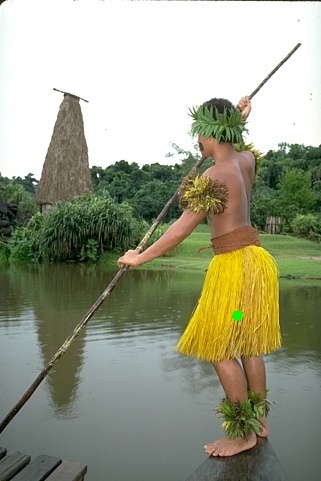}&
			\includegraphics[width=0.14\linewidth,height=3.7cm]{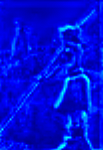}&
			\includegraphics[width=0.14\linewidth,height=3.7cm]{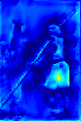}&
			\includegraphics[width=0.14\linewidth,height=3.7cm]{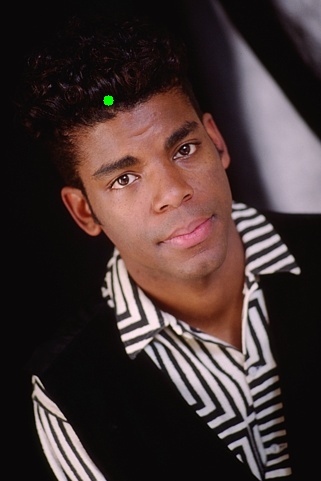}&
			\includegraphics[width=0.14\linewidth,height=3.7cm]{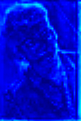}&
			\includegraphics[width=0.14\linewidth,height=3.7cm]{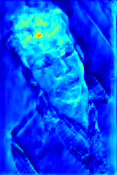}
 \\
		
			\includegraphics[width=0.14\linewidth, height=1.6cm]{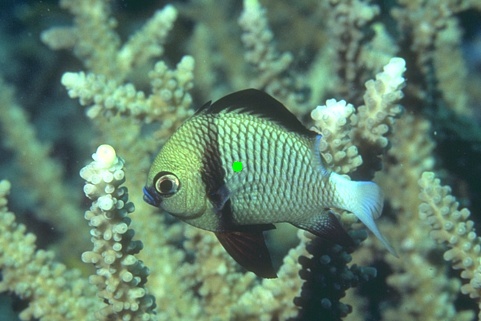}&
			\includegraphics[width=0.14\linewidth, height=1.6cm]{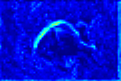}&
			\includegraphics[width=0.14\linewidth, height=1.6cm]{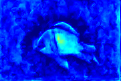}&
			\includegraphics[width=0.14\linewidth, height=1.6cm]{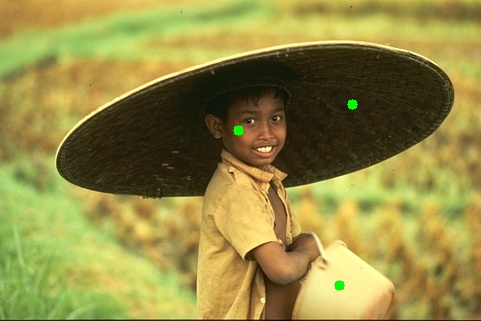}&
			\includegraphics[width=0.14\linewidth, height=1.6cm]{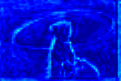}&
			\includegraphics[width=0.14\linewidth, height=1.6cm]{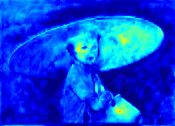}
			 \\
			
			{\scriptsize image and clicks} &
			{\scriptsize model w/o early-late fusion} &
			{\scriptsize model + early-late fusion} &
			{\scriptsize image and clicks} &
			{\scriptsize model w/o early-late fusion} &
			{\scriptsize model + early-late fusion}  \\
			
		\end{tabular}
	\end{center}
    \caption{The qualitative results of early-late fusion. We visualize the feature map before OCR decoder. The second and  fifth column utilize model with early fusion strategy. The third and sixth column utilizes early-late fusion strategy. The architecture of edge-guided flow utilizes early-late fusion strategy and prevents the network forgetting interactive information.} 
    \label{fig:feature_base_vs_edge}
\end{figure*}

\subsection{Network architecture}

\begin{figure*}[t]
	\begin{center}
	    \begin{tabular}{cc}
    		\includegraphics[width=0.49\linewidth]{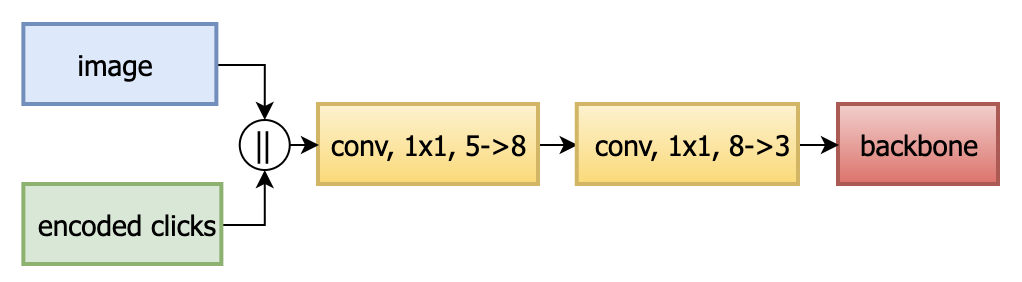} &
    		\includegraphics[width=0.47\linewidth]{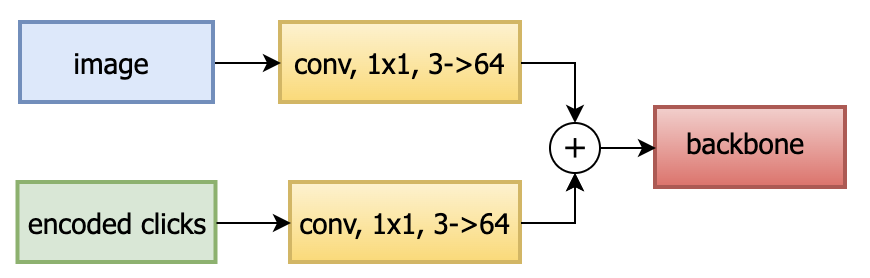}\\
    		{\scriptsize (a)} &
    		{\scriptsize (b)}\\
    	
		\end{tabular}
	\end{center}
	\caption{Different fusion blocks in front of backbone.}
	\vspace{-0.1in}
	\label{fig:encode}
	\vspace{-0.1in}
\end{figure*}

As shown in Fig.~\ref{fig:structure}, the proposed architecture consists of two parts, input base segmentation model (CoarseNet) and refinement model (FineNet). For network inputs, the edge mask and the positive/negative clicks are encoded by the interactive branch. 

\textbf{CoarseNet}. We utilize HRNet-18~\cite{hrnet} as the backbone and OCRNet~\cite{ocr} as segmentation head. The architecture has been proved to be excellent for semantic segmentation tasks. Unlike the single image branch in the generic segmentation model, we add an interaction branch into the backbone to handle the positive clicks, negative clicks, as well as the edge mask. In the beginning, the edge mask is initialized to a zero map with the same size as the input image. After receiving user clicks, the edge mask is estimated through the segmentation result and then is adopt as a part of input. To obtain an accurate edge mask, we add an auxiliary block for the edge constraint in the segmentation head. The fusion details of the image and interaction branch will be introduced in section~\ref{section:edge}.

\textbf{FineNet}. We utilize a FineNet module to further refine the coarse segmentation mask. The module takes three parts as the input, the outputs of CoarseNet, the original image and the user clicks, where the image and the user clicks are the same as partial inputs in CoarseNet. Note that for better visualization, we do not connect them to FineNet in Fig.~\ref{fig:structure}, while they are connected in our implementation. To improve computation efficiency, we use three atrous convolution blocks with lightweight operations in the FineNet. The atrous convolutions also obtain large receptive fields on high-resolution features, so that they can obtain more context information and improve the quality of outputs.

\subsection{Edge-guided Flow}
\label{section:edge}

\textbf{Clicks encoding}: Interactive information contains the coordinates of positive and negative clicks.
The positive clicks emphasize the target objects and the negative clicks isolate abandoned areas. To achieve the intuition and feed to the network, the clicks need to be encoded as feature maps. Usually, interactive segmentation methods utilize Gaussian algorithm or $\mathcal{L}_2$ distance to generate distance maps as the feature maps. However, such distance maps change dramatically when adding a new click. Benenson~\etal~\cite{disk} found that the disk map has more stable performance than other encoding methods. Inspired by~\cite{disk, ritm}, in our implementation, we utilize the disk map with a radius of 5 pixels to encode both positive and negative clicks.

\textbf{Early-late fusion}: In general, a segmentation model only supports three-channel images as the inputs. Thus, the generic model should be modified to adapt to the additional interaction inputs.
Some methods~\cite{f-brs, dextr,itis} introduce additional convolution layers in front of the backbone to accept N-channels input, which is shown in Fig.~\ref{fig:encode}. Both of them fuse interaction and image features before the backbone network, which is so-called early fusion. The early fusion methods have common problems in that interaction information is not extracted properly. The interaction features are much sparser than image features and contain high-level information, e.g. position information. The early layers in the backbone network are focusing on low-level feature extraction, so that the interaction features would be diluted through the early layers and the network can not respond to user clicks timely.

To prevent feature dilution, we propose an early-late fusion strategy to integrate interaction and image features. Instead of only fusing the features at the beginning of the network, we design the multi-stage feature fusion as shown in Fig.~\ref{fig:structure}. The first fusion is the same as Fig.~\ref{fig:encode}(b). The second fusion is between the first transition block and the second stage block of the backbone. The last fusion is after the fourth stage block. The multi-stage fusion promotes the propagation of interaction information over the network and also enables the network to respond to the user clicks precisely.

\textbf{Edge Guidance}: The key idea of interactive segmentation is improving segmentation masks progressively with user clicks. Due to the large spatial variance of the user clicks, the features of consecutive clicks would be quite different, resulting in dramatic segmentation masks. Previous methods~\cite{ritm} introduced the segmentation mask of previous clicks as an input, which alleviates this problem to a certain extent. However, the full mask could make the model fall into local optima, e.g. a poor previous mask usually leads to poor segmentation results. To improve the stability of the segmentation mask, we propose an edge mask scheme, which takes the object edges estimated from the previous iteration as prior information, instead of direct mask estimation. Edge estimation is sparser and less fluctuating than the full mask on input, so it can improve the stability and efficiency of segmentation.

In the interactive segmentation model, the interaction image and edge mask features are heterogeneous, resulting in a large spatial bias. Thus, it is necessary to align these properly. The optical flow method is originally used to align features from two adjacent frames in a video~\cite{flow}. In semantic segmentation, it is effective for multi-scale features alignment while fusing different layers.  Inspired by ~\cite{point_flow,decouple,sfnet}, we take a flow module to align image and interaction features such that spatial information can be represented precisely. The details of the flow module are shown in Fig.~\ref{fig:structure}.

\subsection{Loss Function}

We expect that the loss function is more focusing on the wrong pixels rather than well-classified pixels. Inspired by~\cite{ritm}, we utilize normalize focal loss to calculate the discrepancy between prediction mask and ground truth mask.  It can be denoted as Eq~\ref{equ:dt1}:

\begin{eqnarray}
\label{equ:dt1}
\mathcal{L}_{m}(i, j)=-\frac{1}{\sum_{i, j} (1- p_{i,j})^{\gamma}}(1-p_{i,j})^\gamma\log{p_{i,j}}~,
\end{eqnarray} 
where $\gamma$ is the hyper-parameter for focal loss. $p_{i,j}$ denotes the confidence of the prediction of pixel $(i, j)$, it is denoted as :

 \begin{eqnarray}
 \label{equ:dt3}
p_{i,j}=\begin{cases} p, &y=1\cr 1 - p, & otherwise\end{cases},
\end{eqnarray}
where $p$ is the prediction probability on the location $(i, j)$, $y$ is the corresponding ground truth on the location $(i, j)$.

To minimize the difference between edge estimation from the last iteration and edges derived from ground truth masks, we employ balanced BCE loss $\mathcal{L}_{e}$ to assign more attention on edge than the background. Besides, we utilize BCE as auxiliary loss $\mathcal{L}_{a}$ to constrain the backbone outputs.

\section{Experiments}

\subsection{Experiment Settings}
\begin{table*}[hpt]
	\centering
	
	\caption{Evaluation results of GrabCut, Berkeley, DAVIS and Pascal VOC. Lower value is better. The best result is marked in bold.} 
	\label{tab:sota}
	\vspace{0.1in}
	\begin{tabular}{c|c|c|c|c|c|c} 
		\hline
		\multirow{2}{*}{Method} & \multicolumn{2}{c|}{GrabCut~} & Berkeley~     & \multicolumn{2}{c|}{DAVIS~}   & Pascal VOC     \\ 
		\cline{2-7}
		& NoC@85        & NoC@90        & NoC@90        & NoC@85        & NoC@90        & NoC@85         \\ 
		\hline
		GC~\cite{gc}                      & 7.98          & 10.00         & 14.22         & 15.13         & 17.41         & -              \\
		GM~\cite{gm}                      & 13.32         & 14.57         & 15.96         & 18.59         & 19.50         & -              \\
		RW~\cite{rw}                     & 11.36         & 13.77         & 14.02         & 16.71         & 18.31         & -              \\
		ESC~\cite{esc}                     & 7.24          & 9.20          & 12.11         & 15.41         & 17.70         & -              \\
		GSC~\cite{esc}                     & 7.10          & 9.12          & 12.57         & 15.35         & 17.52         & -              \\ 
		\hline
		DOS~\cite{deep1}            & -             & 6.04          & 8.65          & -             & -             & 6.88           \\
		LD~\cite{ld}        & 3.20          & 4.79          & -             & 5.05          & 9.57          & -              \\
		RIS-Net~\cite{ris}                 & -             & 5.00          & 6.03          & -             & -             & 5.12           \\
		ITIS~\cite{itis}                    & -             & 5.60          & -             & -             & -             & 3.80           \\
		CAG~\cite{cag}                     & -             & 3.58          & 5.60          & -             & -             & 3.62           \\
		BRS~\cite{brs}                     & 2.60          & 3.60          & 5.08          & 5.58          & 8.24          & -              \\
		FCA~\cite{fca}                 & -             & 2.08          & 3.92          & -             & 7.57          & 2.69           \\
		IA+SA~\cite{eccv2020}                   & -             & 3.07          & 4.94          & 5.16          & -             & 3.18           \\
		f-BRS-B~\cite{f-brs}                 & 2.50          & 2.98          & 4.34          & 5.39          & 7.81          & -              \\
		RITM-H18~\cite{ritm}                & \textbf{1.54} & \textbf{1.70} & 2.48          & 4.79          & 6.00          & 2.59           \\ 
		\hline
		Our method                & 1.60          & 1.72          & \textbf{2.40} & \textbf{4.54} & \textbf{5.77} & \textbf{2.50}  \\
		\hline
		
	\end{tabular}
	\vspace{-0.15in}
\end{table*}

\begin{figure*}[!t]
	\begin{center}
		\begin{tabular}{ccc}
			
			\includegraphics[width=0.31\linewidth, height=3.7cm]{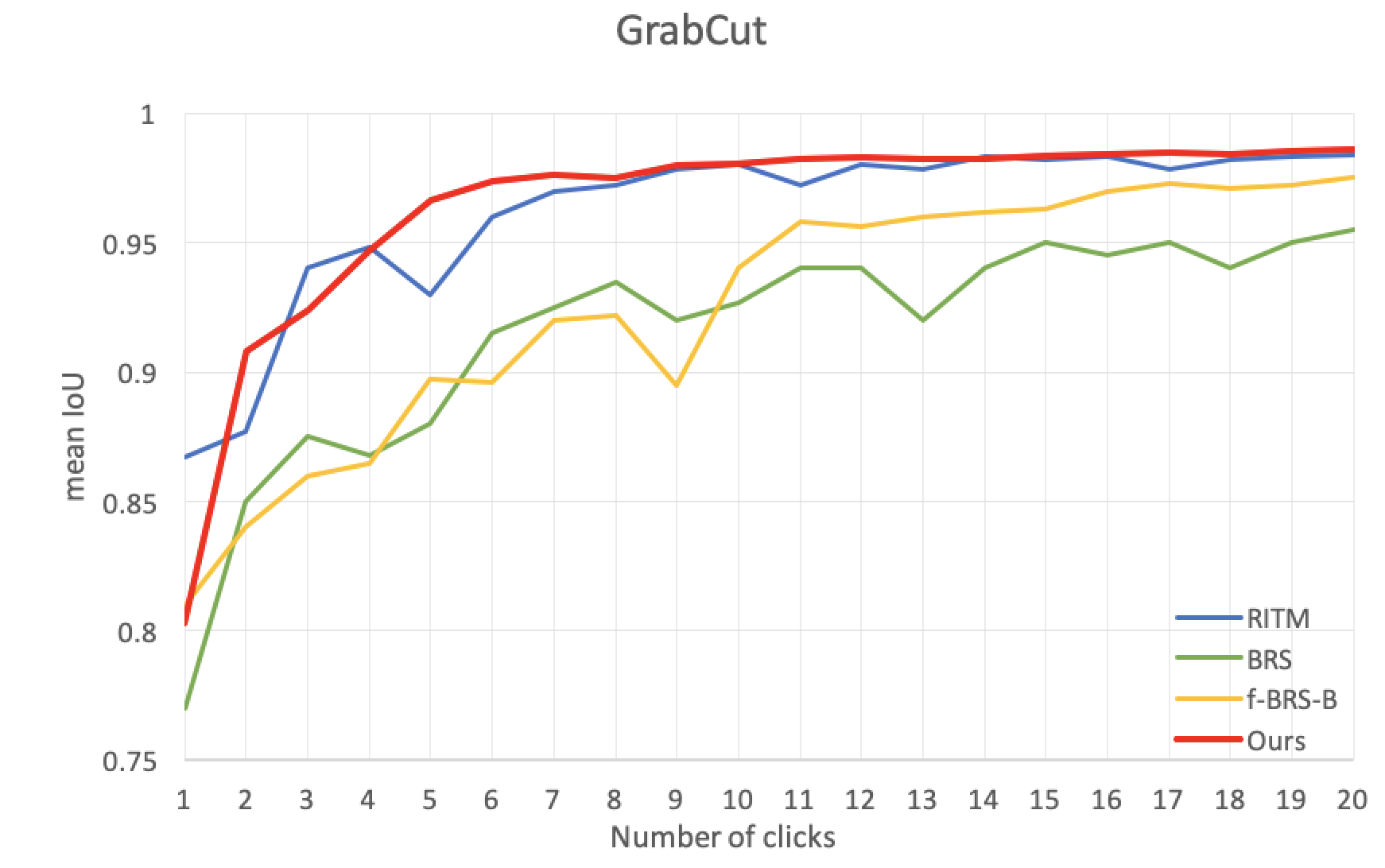}&
			\includegraphics[width=0.31\linewidth, height=3.8cm]{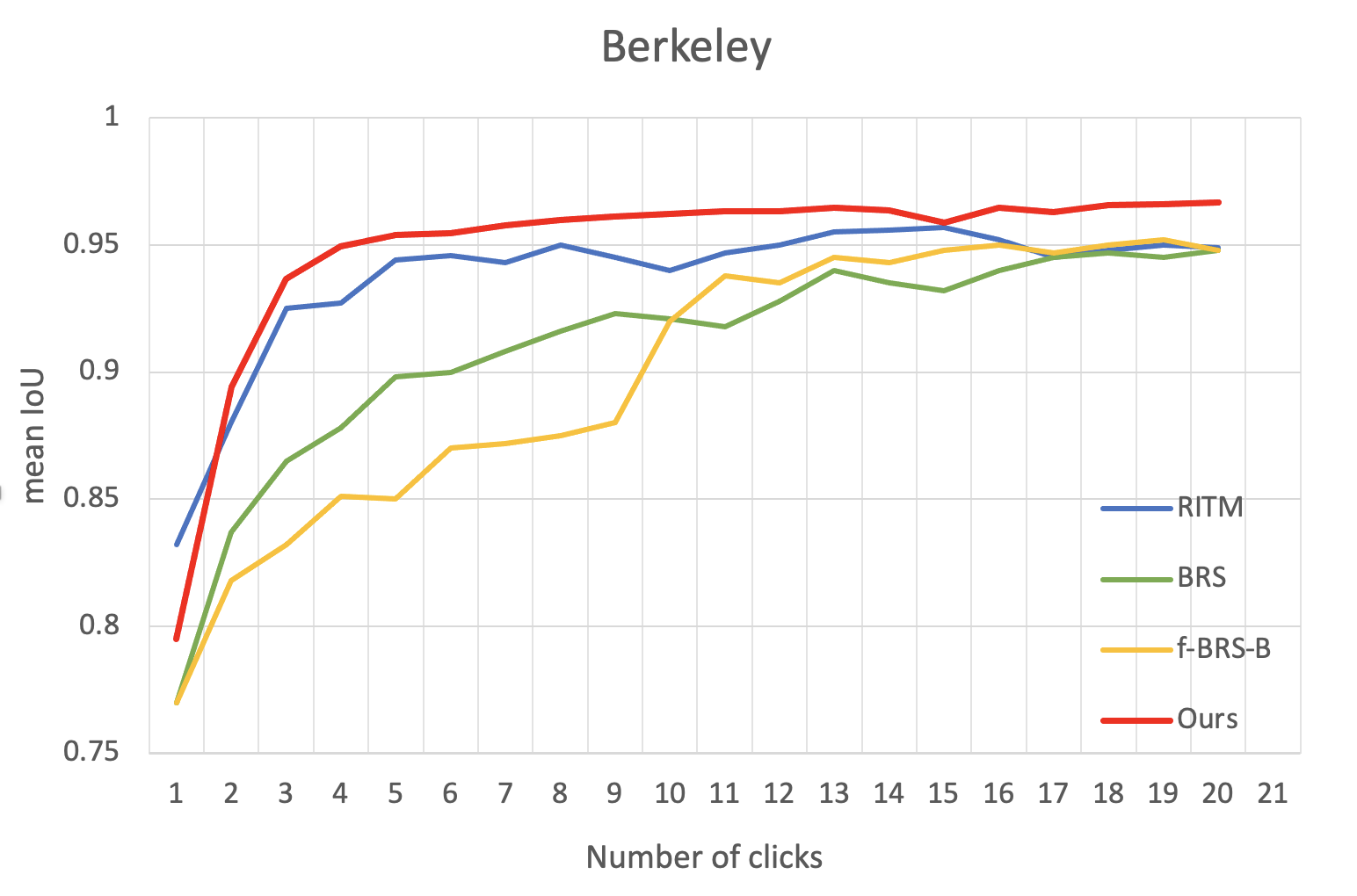}&
			\includegraphics[width=0.31\linewidth, height=3.75cm]{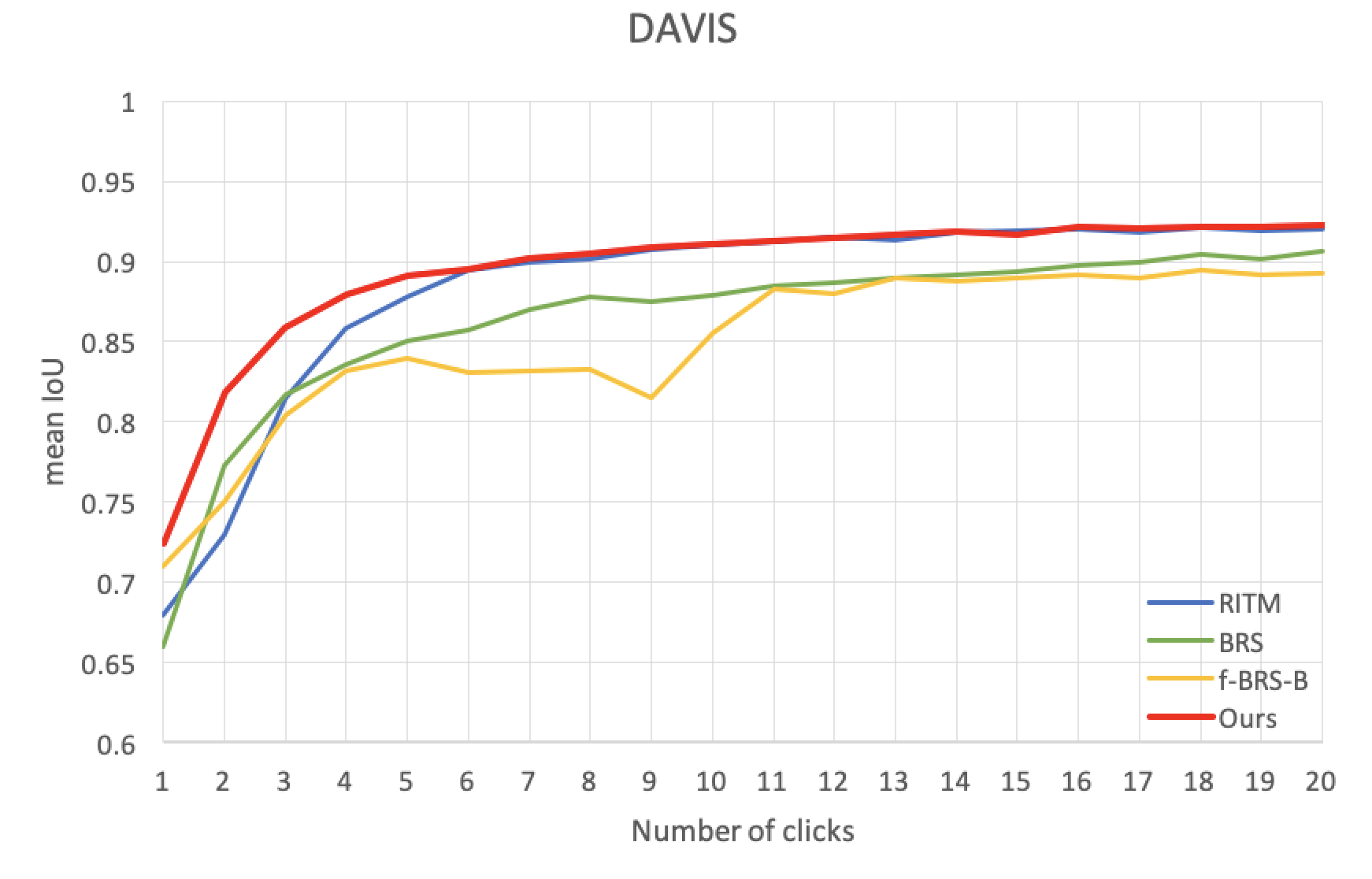}\\
			
		\end{tabular}
	\end{center}
	
    \caption{Evaluation of different interactive segmentation methods.  The plots represent the mean IoU of different methods for the first 20 clicks iteratively on GrabCut, Berkeley and DAVIS.} 
    \label{figshow}
    \vspace{-0.15in}
\end{figure*}

\textbf{Datasets}: Following~\cite{ritm}, we use the combination of COCO~\cite{coco} and LVIS~\cite{lvis} as the training data. LVIS contains 164k images and 2M high-quality instance masks with over 1000 classes. Due to the long-tailed object classes from LVIS dataset, following~\cite{ritm}, we use 10582 COCO images and corresponding 25832 instance masks as the augmented dataset.

We evaluate our method on four popular benchmarks, including GrabCut~\cite{grabcut}, Berkeley~\cite{berkeley}, Pascal VOC~\cite{pascalvoc}, DAVIS~\cite{davis}. The GrabCut dataset contains 50 images and 50 masks. 
The Berkeley dataset contains 96 images and 100 masks.  
For Pascal VOC dataset, we only use the validation set, which contains 1449 images and 3417 instances. 
The DAVIS dataset is randomly sampled from the video object segmentation dataset. We adopt 345 images and corresponding 345 masks which were introduced in~\cite{brs,ritm}.

\textbf{Evaluation metrics}: Mean intersection over union (IoU) and the standard number of clicks (NoC) are used as evaluation metrics. NoC indicates the number of interactive clicks for achieving over the specified IoU threshold, which is commonly set as 85\% and 90\%~\cite{ritm, dextr, f-brs, brs}, i.e. NoC@85 and NoC@90. During the evaluation procedure, we build a simulation to generate clicks, following ~\cite{f-brs, brs, ritm}. The first click is positive to the object of interest. The next click is selected as the centre of the largest error region. Note that in this work, the maximum number of clicks is 20, which is the same as~\cite{f-brs, brs, ritm}. 
\begin{table*}[!t]
	\centering
	\caption{ Evaluation of different model architecture on Berkeley, DAVIS and Pascal VOC.  `baseline' represents coarse model without edge-guided flow branch. `EF' represents edge-guided flow. `F' represents FineNet. Lower value means less interaction and better performance.}
	\label{tab:ablation}
	\begin{tabular}{c|c|c|c|c|c|c}
		\hline
		\multirow{2}{*}{model}      & \multicolumn{2}{c|}{Berkeley} & \multicolumn{2}{c|}{DAVIS} & \multicolumn{2}{c}{Pascal VOC} \\ \cline{2-7} 
		& NoC@85        & NoC@90        & NoC@85       & NoC@90      & NoC@85         & NoC@90        \\ \hline
		baseline                        & 2.06          & 2.93          & 4.83         & 6.27        & 2.68           & 3.26          \\ 
		baseline+EF                   & 1.83          & 2.79          & 4.82         & 6.14        & 2.66           & 3.19          \\ 
		baseline+EF+F       & \textbf{1.60}          & \textbf{2.40}          &\textbf{4.54}        & \textbf{5.77}        & \textbf{2.50}           & \textbf{3.07}          \\ \hline		
	\end{tabular}
\end{table*}

\begin{table*}[!t]
	\centering
	\caption{Evaluation of different input setting on Berkeley, DAVIS and Pascal VOC. `SI' represents sobel edge as prior information. `EI' represents edge mask from segmentation result. `F' represents FineNet. Lower value means less interaction and better performance.}
	\label{tab:abl2}
		\begin{tabular}{c|c|c|c|c|cc}
			\hline
			\multirow{2}{*}{Model}         & \multicolumn{2}{c|}{Berkeley} & \multicolumn{2}{c|}{DAVIS}    & \multicolumn{2}{c}{Pascal VOC}                     \\ \cline{2-7} 
			        & NoC@85        & NoC@90        & NoC@85        & NoC@90        & \multicolumn{1}{c|}{NoC@85}       & NoC@90        \\ \hline
			        
			SI + w/o F         & 1.83          & 3.20          & 5.04          & 6.32          & \multicolumn{1}{c|}{2.88}         & 3.41          \\
			EI + w/o F               & 1.83          & 2.79          & 4.82          & 6.14          & \multicolumn{1}{c|}{2.66}         & 3.19          \\
			SI + F            & \textbf{1.58} & 2.73          & 4.68          & 6.21          & \multicolumn{1}{c|}{2.9}          & 3.46          \\
			EI + F    & 1.60          & \textbf{2.40} & \textbf{4.54} & \textbf{5.77} & \multicolumn{1}{c|}{\textbf{2.5}} & \textbf{3.07} \\ \hline
	\end{tabular}
	\vspace{-0.1in}
\end{table*}

\textbf{Implementation details}: In this work, we utilize COCO+LVIS dataset to train our model. For data augmentation, we adopt random resize with a scale from 0.75 to 1.4, horizontal flip, random crop image with the resize of (320, 480). For colour distortion, we utilize random distribution on contrast, brightness and pixel value for the original image.

In our experiments, the weight of focal loss $\mathcal{L}_{m}$, balanced BCE loss $\mathcal{L}_{e}$ and BCE loss $\mathcal{L}_{a}$ is set as 1, 0.4, 0.4, respectively. Firstly, we train the CoarseNet for 40 epochs with the learning rate of $5\times10^{-4}$. Then, we train the whole network for the last 30 epochs, where the learning rate of $5\times10^{-4}$ for FineNet and $5\times10^{-5}$ for CoarseNet. We utilize Adam optimizer and polynorm decay to reduce the learning rate to 1 percent of the initial learning rate value. Our model is based on PaddlePaddle~\cite{paddlepaddle, paddlepaddle2}. We train the model on 2 GPUs (Tesla V100), and the batch size is set to 32. 

\subsection{Comparison with SOTA methods}

\textbf{NoC@k}.We calculate NoC@85 and NoC@90 as the metrics
on GrabCut, Berkeley, DAVIS and Pascal VOC.  The evaluation results are shown in table~\ref{tab:sota}. Our method achieves the best performance on all datasets except for GrabCut. GrabCut has a limited number of objects with clear boundaries, so that all deep learning methods perform very well, and our method also achieves competitive results. Whereas Berkeley images have complicated object boundaries, e.g. bicycle wheel springs and parachute ropes.
In this dataset, our method achieves the best performance compared with other methods.  Since our method utilizes edge masks as prior information, it provides more accurate details of segmentation objects. 

For DAVIS and Pascal VOC datasets, they contain more than 1.5k images with various scenes, which are more suitable for practical tasks. Our method achieves the best results compared with other methods. Analysing the result, we consider that early-late fusion strategy can prevent features of interactive clicks dilution over the network. Therefore, our method responds to the clicks efficiently and is robust in various scenes.

\textbf{mIoU per click.} Fig.~\ref{figshow} shows the mIoU change over the first 20 clicks on different datasets. There are three observations: 1) Our method has the most stable performance compared with other methods, where mIoU improves gradually as the clicks increases. However, other methods have the degradation problem. For example, mIoU of RITM decreases dramatically when the clicks increase from 4 to 5, and 10 to 11 on GrabCut dataset. As described above, the proposed edge estimation can improve the segmentation stability. 2) Our method requires fewer clicks to achieve higher mIoU compared with other methods. In the Berkeley dataset, NoC@95 of our method is around 4, while others are more than 10.  3) Using all 20 clicks, our method still achieves the highest mIoU on all datasets. The experiment demonstrates the superior segmentation ability of our method.
\begin{figure*}[!t]
	
	\begin{center}
		\begin{tabular}{ccc}

			\includegraphics[width=0.31\linewidth, height=3.8cm]{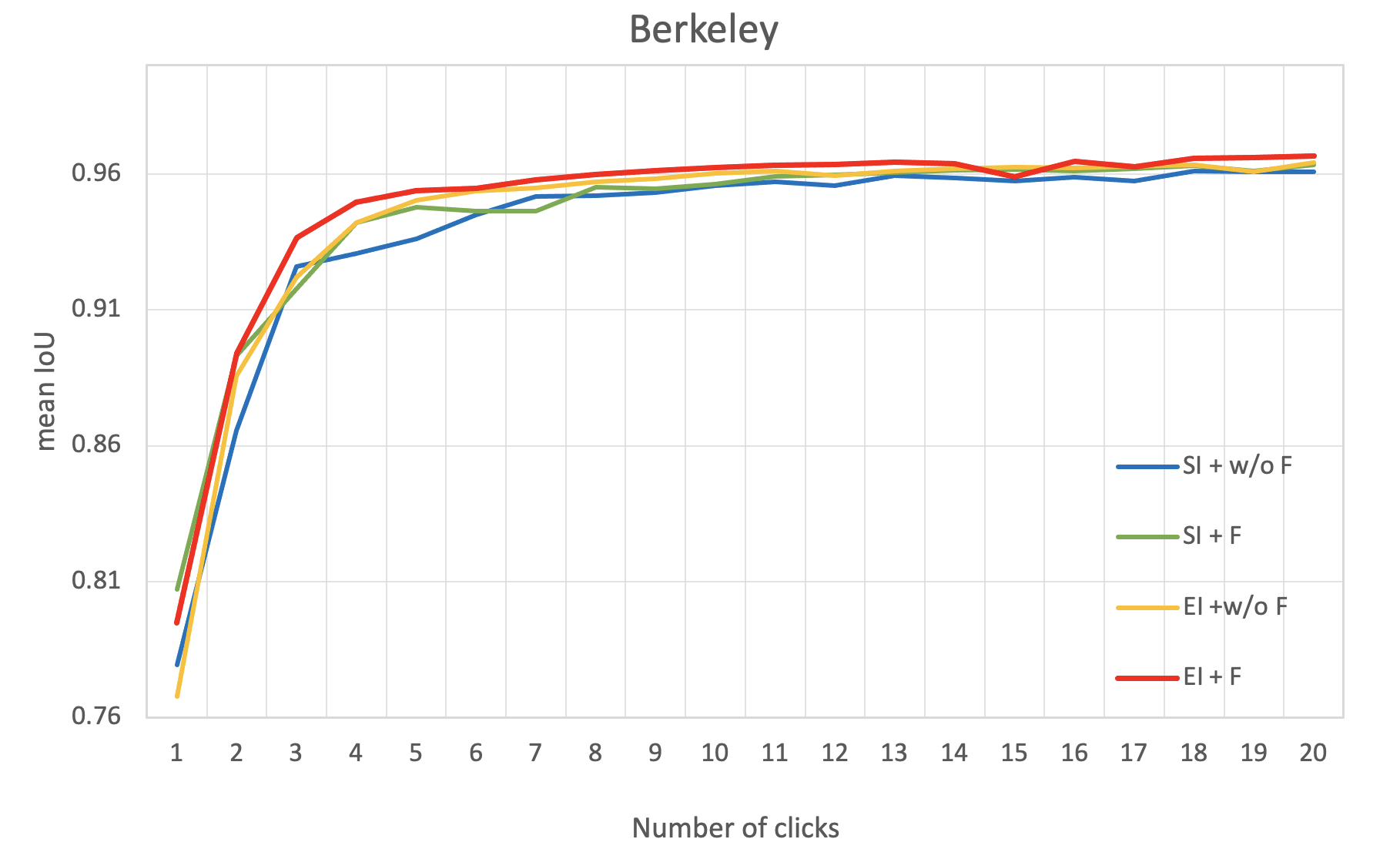}&
			\includegraphics[width=0.31\linewidth, height=3.8cm]{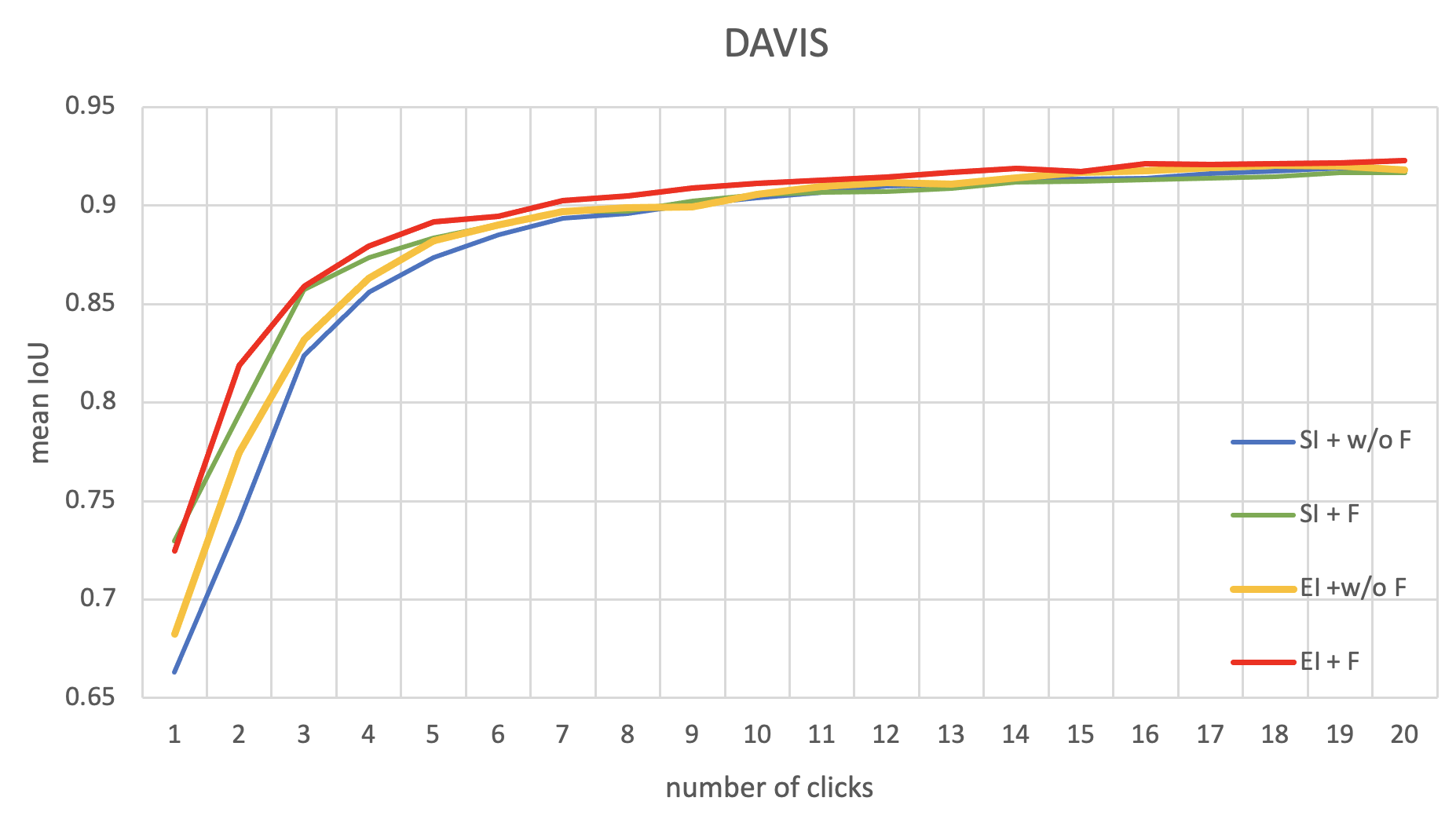}&
			\includegraphics[width=0.31\linewidth, height=3.8cm]{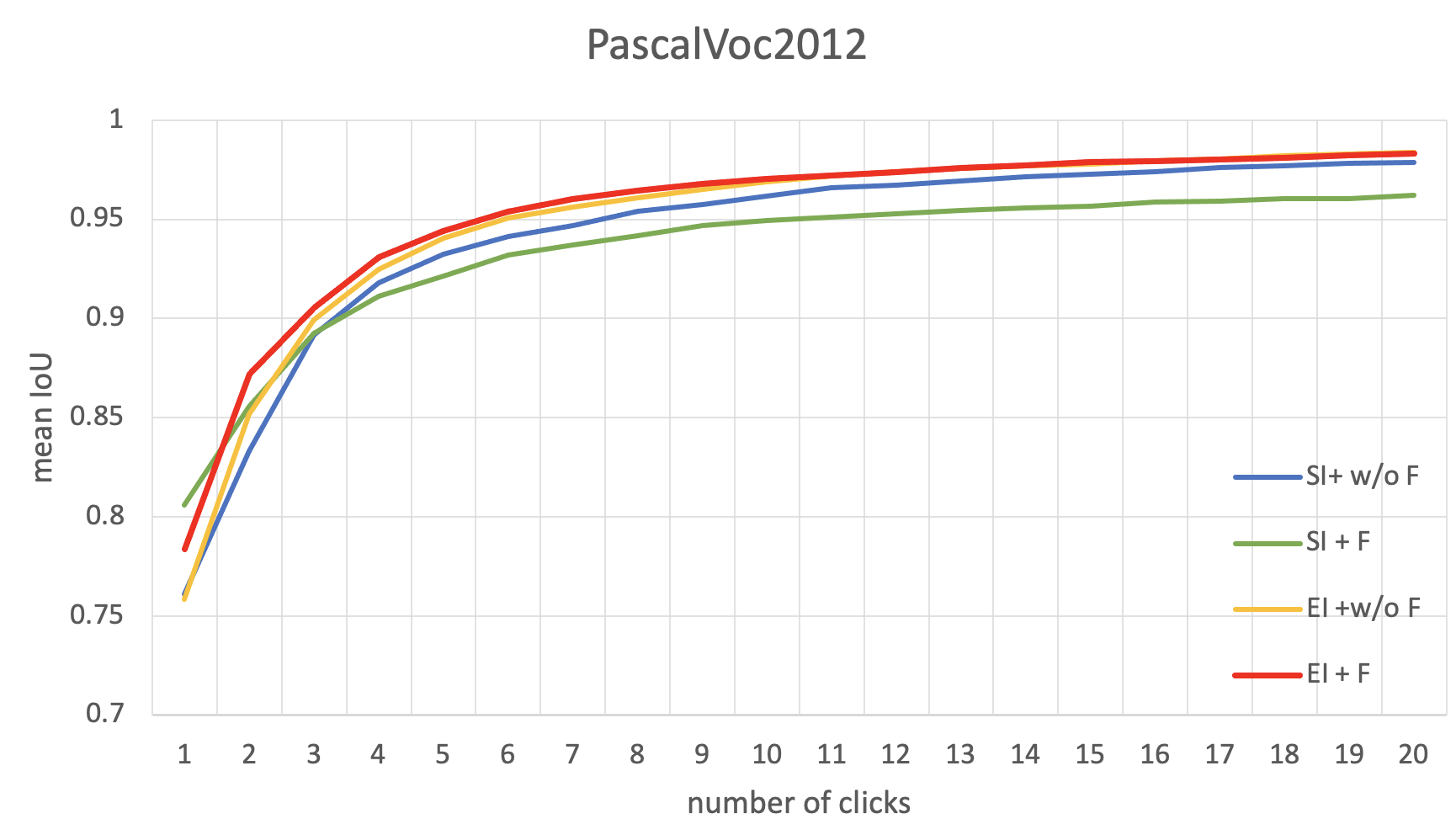}\\
			
		\end{tabular}
	\end{center}
    \caption{Effects of prior information.  The plot represents the value of mean Intersection over Union for first 20 clicks iteratively on Berkeley, DAVIS and  Pascal VOC, respectively.} 
    \label{figimg2}
    \vspace{-0.1in}
\end{figure*}

\begin{figure*}[th]
	\setlength{\abovecaptionskip}{0.cm}
	\setlength{\belowcaptionskip}{-0.cm}
	\begin{center}
		\begin{tabular}{ccc}
			
			\includegraphics[width=0.31\linewidth]{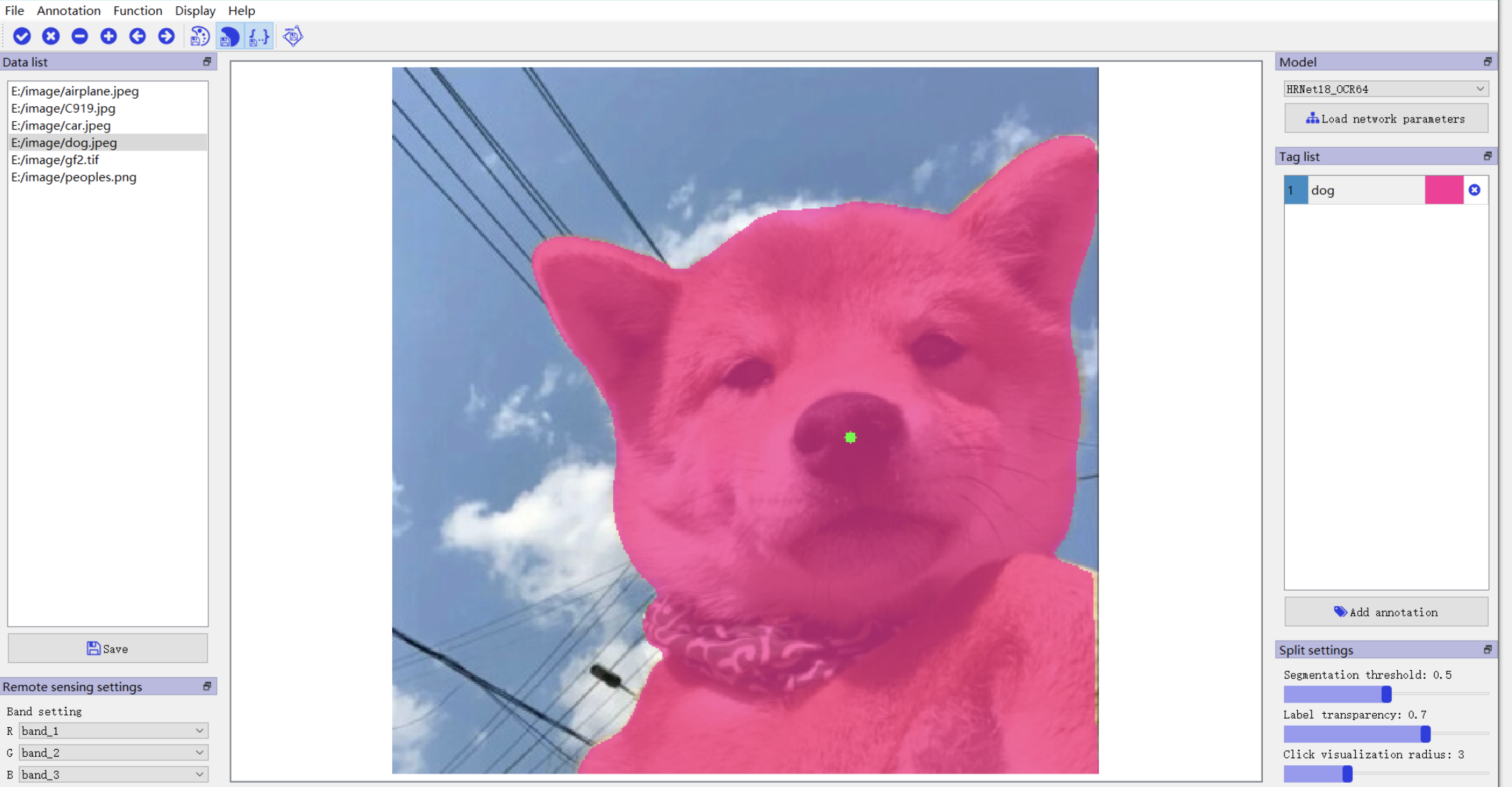}&
			\includegraphics[width=0.31\linewidth]{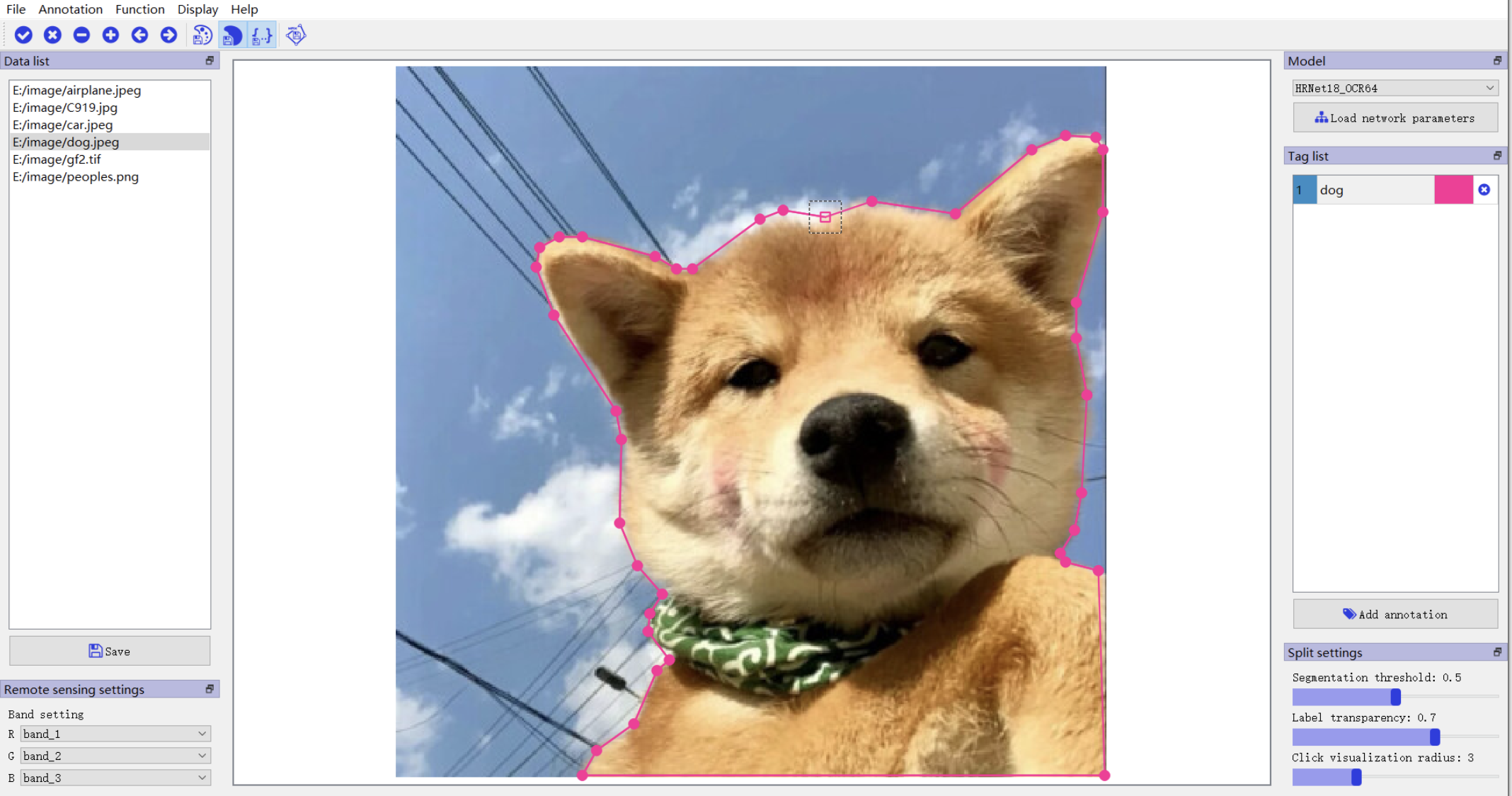}&

			\includegraphics[width=0.31\linewidth]{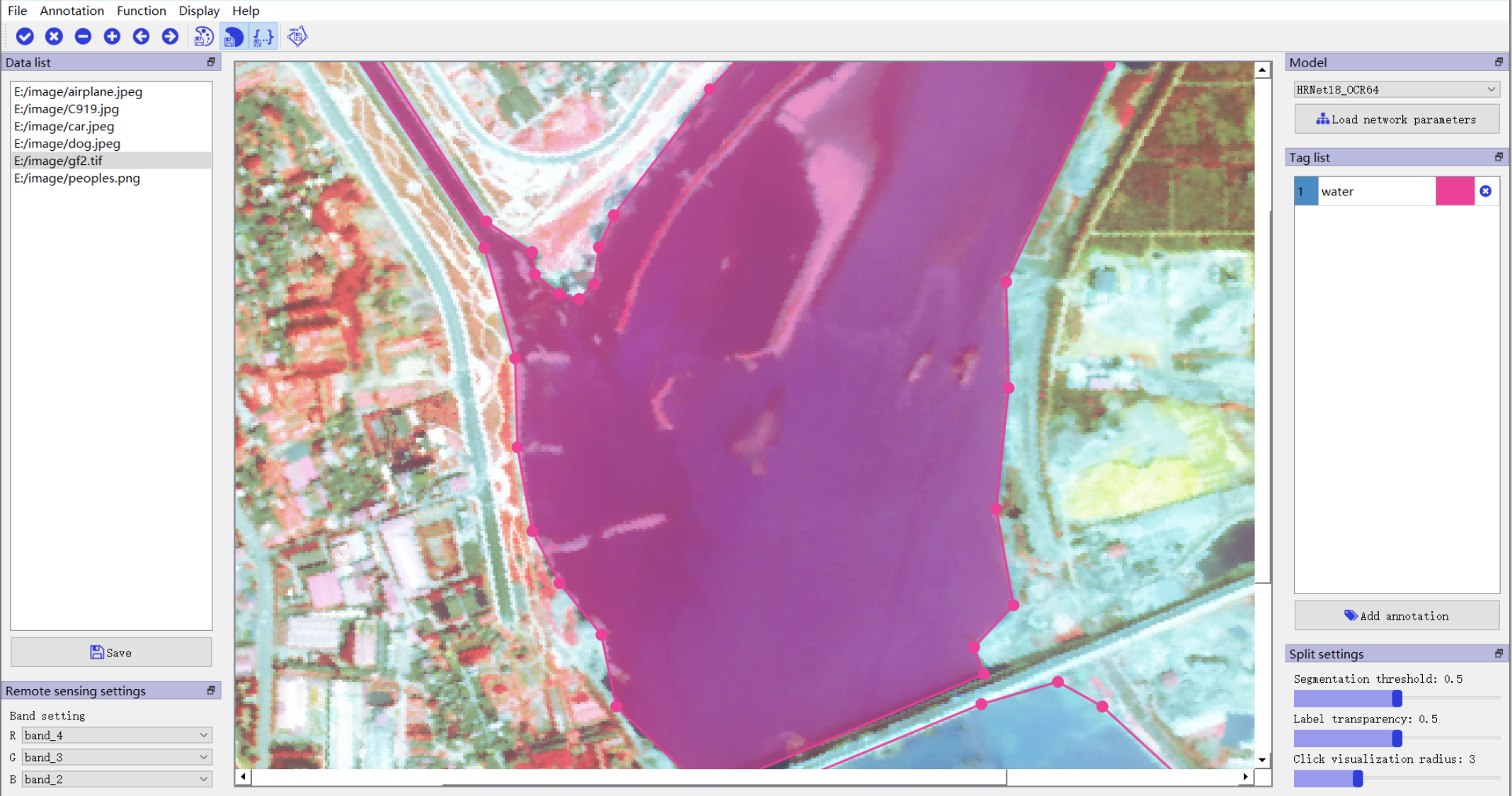}\\

			{\scriptsize (a) UI for interactive clicks} &
			{\scriptsize (b) adjustment process by polygon editing}&
			{\scriptsize (c) application on remote sense} \\
			
		\end{tabular}
	\end{center}
	\vspace{-0.05in}
    \caption{Visualization of interactive segmentation tool.} 
    \label{fig:ui}
    \vspace{-0.15in}
\end{figure*}

\subsection{Ablation Study}

\textbf{Edge Flow}. We perform ablation experiments on Berkeley, DAVIS and Pascal VOC datasets to validate the effectiveness of the edge-guided flow. As shown in table~\ref{tab:ablation}, the baseline with edge-guided flow outperforms the baseline obviously on all three datasets. For a better understanding, we visualize the intermediate feature maps of baseline and baseline+EF, as shown in Fig.~\ref{fig:feature_base_vs_edge}. The model with the early-late fusion shows the stronger activation on coordinate information, while the early fusion has the feature dilution problem. Then, we visualize the output difference between baseline and baseline+EF, which is shown in Fig.~\ref{fig:final_base_vs_edge}. The result demonstrates that edge-guided flow improves interactive segmentation performance and stabilize the output when adding a new click. We apply edge masks as prior information of the network, which makes the segmentation results more stable.

\textbf{FineNet}. As shown in table~\ref{tab:ablation}, we find that NoC@85 and NoC@90 improve significantly after FineNet is added. As described above, FineNet adopts atrous convolution blocks which obtain large receptive fields on high-resolution features, so that they can obtain more context information. The best result is the combination of baseline+EF+F, which demonstrates the effectiveness of the early-late fusion strategy and the coarse-to-fine architecture.

\textbf{Prior information} 
Table~\ref{tab:abl2} shows the effect of prior information on different datasets. Fig~\ref{figimg2} shows the mIoU change over the first 20 clicks using different prior information. We study two types of edge masks, 1) the edge mask estimated from the previous segmentation result, i.e. EI mask, 2) the edge mask directly calculated by the Sobel operator, i.e. SI mask. We find the model with the EI mask outperforms the other one.
In the segmentation model, prior information can be dynamically adjusted with the network and user interaction clicks. Therefore, the EI mask is more related to the object of interest. However, the SI mask calculation is only based on the original image, where it contains a lot of redundant information which is irrelevant to the object of interest. Therefore, it causes network performance degradation. 

\section{Interactive Segmentation Tool}

In this work, we develop an interactive segmentation tool using the proposed model. Our tool aims to help users annotate segmentation datasets efficiently and accurately. The annotation pipeline includes three steps: data preparation, interactive annotation and polygon frame editing.

During data preparation, our tool supports data formats in multiple domains including natural images, medical imaging, remote sensing images and so on. After the image is loaded into the software, it can be zoomed, moved and pre-processed by adjusting brightness and contrast. The annotation process is highly flexible and configurable. Most operations in the application support keyboard shortcuts which can be changed by the user.

During interactive annotation, users add positive and negative points with left and right mouse clicks, respectively. The application runs model inference and shows the user prediction result, as shown in Fig.~\ref{fig:ui}(a). The user can adjust the target border by changing the threshold to distinguish foreground and background pixels to get more accurate segmentation results. The tool also supports filtering the largest connected region, which is shown in Fig.~\ref{fig:ui}(b). This feature is useful when there are multiple targets of the same type in an image. Suppressing small positive regions can free users from clicking negative points in each of these regions.

After finishing interactive segmentation, the tool generates a polygon frame around the target border. Users can adjust the polygon vertexes to further improve segmentation accuracy. It is flexible for practical segmentation tasks. In some cases, adjusting the polygon frame could be faster than adding many clicks during interactive segmentation, so that it improves the overall annotation efficiency. Finally, the segmentation results can be saved in multiple formats including segmentation mask, PASCAL VOC, COCO, pseudo colour and so on.

\section{Conclusion}

A large number of labelled image data are usually essential for segmentation models. Due to the high cost of the pixel-level annotations, interactive segmentation becomes an efficient way to extract the object of interest. In this work, we propose a novel interactive architecture named EdgeFlow that fully utilizes the user interaction information without any post-processing or iterative optimization scheme. With the coarse-to-fine network design, our proposed method achieves state-of-the-art performance on common benchmarks. Furthermore, we develop an efficient interactive segmentation tool that helps the user to improve the segmentation result progressively with flexible options. In future work, we will work on lightweight models that can be deployed into various platforms. Another promising topic we are working on is multi-modality, which utilizes more input types, like audio and text. The different kinds of inputs can complement each other, so that it is vital to improve the quality further.

\section{Acknowledgements}

This work was supported by the National Key Research and Development Project of China (2020AAA0103500).

{\small
\bibliographystyle{ieee_fullname}
\bibliography{egbib}
}

\end{document}